\documentclass[conference]{IEEEtran}
\usepackage{cite}
\usepackage{amsmath,amssymb,amsfonts}
\usepackage{algorithmic}
\usepackage{textcomp}
\usepackage{xcolor}
\def\BibTeX{{\rm B\kern-.05em{\sc i\kern-.025em b}\kern-.08em
    T\kern-.1667em\lower.7ex\hbox{E}\kern-.125emX}}

\usepackage[pdftex]{graphicx}
\graphicspath{{images/}}
\DeclareGraphicsExtensions{.jpeg,.jpg,.png}
\usepackage{mathptmx} 
\usepackage{times} 
\usepackage{amsmath} 
\usepackage{amssymb}  
\usepackage{hyperref}
\usepackage{cleveref}
\usepackage{hyperref}

\begin{document}

\title{Deployment of Aerial Robots during the Flood Disaster in Erftstadt / Blessem in July 2021}

\author{\IEEEauthorblockN{Hartmut Surmann}
\IEEEauthorblockA{\textit{Computer Science Department} \\
\textit{University of Applied Science}\\
Gelsenkirchen, Germany \\
hartmut.surmann@w-hs.de}
\and
\IEEEauthorblockN{Dominik Slomma, Robert Grafe}
\IEEEauthorblockA{\textit{German Rescue Robotic Centre} \\
Dortmund, Germany \\
Dominik.10041991@web.de \\
robert.grafe@rettungsrobotik.de}
\and
\IEEEauthorblockN{Stefan Grobelny}
\IEEEauthorblockA{\textit{Institute of Fire Service \&} \\
\textit{Rescue Technology,}\\
 Fire Department of Dortmund, Germany \\
grobelny@consatex.de}
}


\maketitle
\begin{abstract}
Climate change is leading to more and more extreme weather events such as heavy rainfall and flooding. This technical report deals with the question of how rescue commanders can be better and faster provided with current information during flood disasters using Unmanned Aerial Vehicles (UAVs), i.e. during the flood in July 2021 in Central Europe, more specifically in Erftstadt / Blessem. The UAVs were used for live observation and regular inspections of the flood edge on the one hand, and on the other hand for the systematic data acquisition in order to calculate 3D models using Structure from Motion and MultiView Stereo. The 3D models embedded in a GIS application serve as a planning basis for the systematic exploration and decision support for the deployment of additional smaller UAVs but also rescue forces. The systematic data acquisition of the UAVs by means of autonomous meander flights provides high-resolution images which are computed to a georeferenced 3D model of the surrounding area within 15 minutes in a specially equipped robotic command vehicle (RobLW). From the comparison of high-resolution elevation profiles extracted from the 3D model on successive days, changes in the water level become visible. This information enables the emergency management to plan further inspections of the buildings and to search for missing persons on site.
\end{abstract}

\begin{IEEEkeywords}
Rescue Robotic, UAVs
\end{IEEEkeywords}

\section{Introduction}
\label{introduction}
\begin{figure}[!t]
\centering
\includegraphics[width=0.4\textwidth]{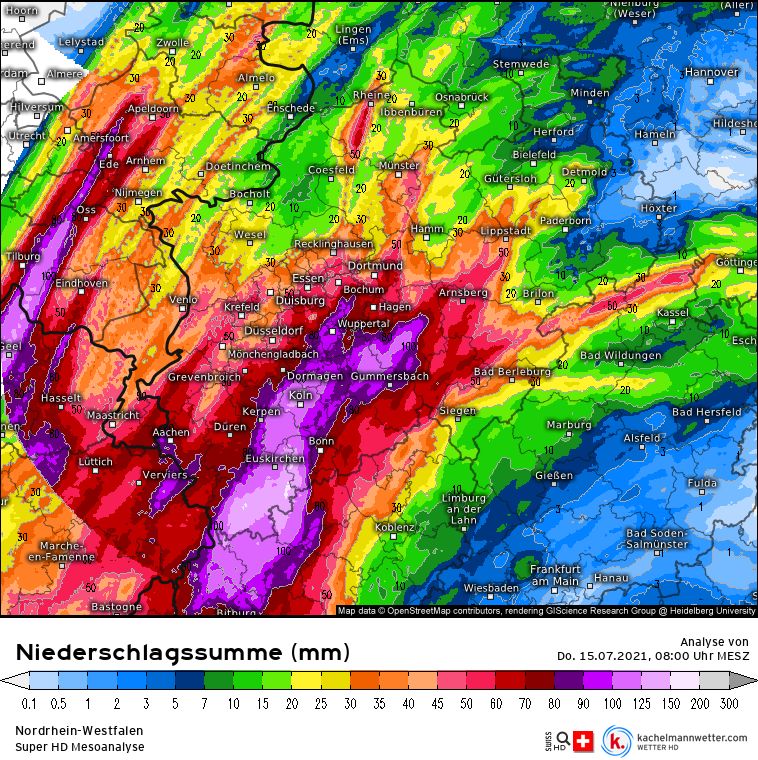}
\caption{Total rainfall in mm at Thursday, July 15, 8pm  }
\label{fig:rainfall}
\end{figure}

The Flood in Western and Central Europe in July 2021 was a natural disaster with severe flash floods in several river basins in Central Europe in the summer of 2021 (Fig. \ref{fig:rainfall}\footnote{source from \url{kachelmannwetter.com}}). Parts of Belgium, the Netherlands, Austria, Switzerland, Germany and other neighboring countries were particularly affected. The most severe floods were caused by thunderstorm "Bernd" \footnote{2021 European floods \url{https://en.wikipedia.org/wiki/2021_European_floods}}. 

The drastic consequences of the storm disaster in Western Germany also made themselves strongly felt in Erftstadt / Blessem. Due to the flooding and a potential collapse of the "Steinbach" dam,  thousand residents in several localities had to be evacuated from their homes. The Erft and Swist rivers which had burst their banks, flooded large parts of the Erftstadt urban area and long-distance roads such as the federal highways No. 1 (Eifelautobahn) and No. 61 as well as the federal highway 265 were closed as a result of the flooding and road damage. In the Erftstadt district of Blessem, the waters of the Erft flowed through a residential and commercial area and made a new path into the pit of the Blessem gravel plant; several houses were washed out, several others damaged near Blessem Castle.

As a result, an extensive emergency response mission commenced, including a rescue robotics team of the German Rescue Robotic Center (DRZ). The team consisted of personnel from Dortmund fire department (FwDO), the DRZ, University of Bonn and Lübeck, TU Darmstadt and the Westphalian University of Applied Science (WHS) and set off to Erftstadt/Blessem for a two-day mission with the RobLW directly in this very dangerous area at Blessem Castle. The team had four tasks:
\begin{enumerate}
    \item Secure the emergency responders during their missions in the danger zone, especially during the search of the city's buildings for missing persons. 
    \item Generate high-resolution 3D models for the purpose of further mission planning directly on site even without power, internet and mobile phone connection. 
    \item Detailed inspection of all buildings / structures that cannot be accessed or reached.
    \item Creating clear and easily accessible documentation for the emergency services.
\end{enumerate}



\section{State of the art}
\label{sec:sta}
Rescue robotics is attracting more and more attention in practice, so it is fundamentally important in this area to test the robots as often as possible in real operations and to implement the knowledge gained directly \cite{app11010363}. Thus, in February 2021 in Berlin, after an industrial fire with a conventional DJI UAV, a standard panoramic camera and appropriate software, information could be collected to determine the cause of the fire \cite{9597677}.  Ground robots could not be used there because the ground was not passable due to contaminated extinguishing water and the building was also in danger of collapsing. The fact that ground robots cannot be used optimally in such areas was demonstrated in Cologne in 2009 \cite{5981550}, where a collapsed building was to be explored by a ground robot, but debris prevented it from penetrating certain areas. The same problems occurred in Japan in 2011 after a major earthquake, where the robots were to search for missing persons \cite{matsuno2014utilization}. Therefore, in addition to ground robots, underwater robots were also used. After strong earthquakes in 2012 and 2016 in Italy, ground, air and water robots were used to investigate building damage as they threatened to collapse \cite{Kruijff-amatrice, advanced-robotics-2014}. 
Since the robots are supposed to provide optimal support to the responders, it is extremely important to train the responders and test and optimize the robots in various real-world environments \cite{7017681}. Fukushima for example provides an optimal testing ground for the real-life use of robots \cite{9419563}. At the DRZ, various robot platforms are developed and then evaluated and validated in the associated Living Lab \cite{9597869}. This provides another opportunity to test and optimize the robots under conditions that are as real as possible. 
In addition to the missions and the test areas, the use of UAVs in the rescue area is shown to play a fundamental role, as they can explore a large area in a short time. Thus, conclusions can be drawn more quickly and faster action is possible \cite{mayer:hal-02128385, surmann2019ssrr}. Online or offline 3D reconstruction / mapping (Aerial photogrammetry) from multiple images can be done with, e.g., VisualSFM, COLMAP, OpenMVS, OpenDroneMap, MVE, DroneDeploy, Pix4D Mapper, AutoDesk, Agisoft PhotoScan \cite{Opendronemap,Fuhrmann:2014:MMR:2854922.2854925}
or online with, e.g., LDSO, ORB2 Slam or SVO / REMODE \cite{gao2018ldso,murTRO2015,Pizzoli2014ICRA}.

Some European projects address the deployment of (teams of) UGVs and UAVs in various disaster response scenarios.
ICARUS \cite{icarus} and DARIUS \cite{darius} target the development of robotic tools that can assist during disaster response operations
focusing on autonomy.
SHERPA \cite{sherpa} is focused on the development of ground and aerial  robots  to  support  human-robot team response in an alpine scenario.

\begin{figure}[!t]
\centering
\includegraphics[width=0.45\textwidth]{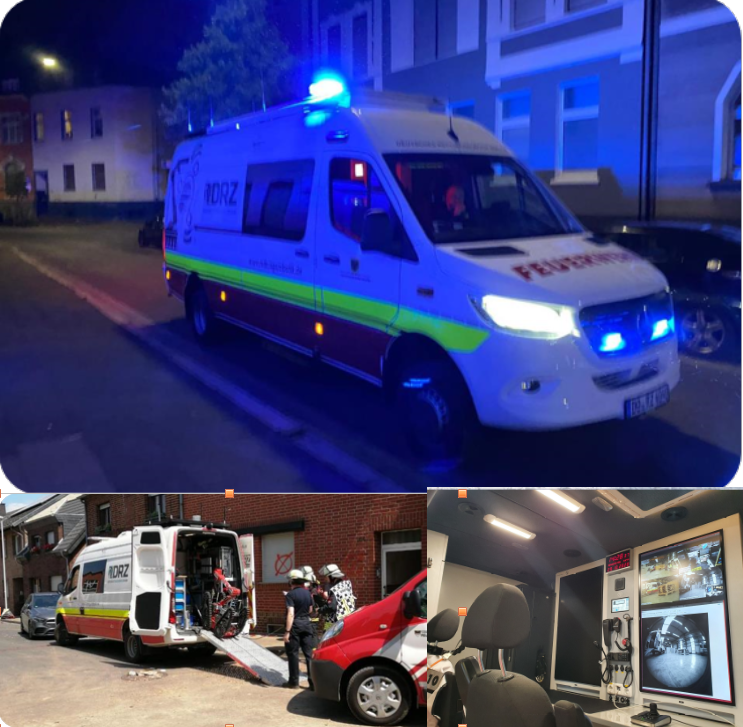}
\caption{Overview of the RobLW. Top: The RobLW. Bottom left: Trunk with a ground robot. Bottom right: The command center with two monitor workstations.}
\label{fig:rlw-overview}
\end{figure}

\section{Deployment}
\subsection{Hardware and software equipment}
\subsubsection{RobLW}
The RobLW (fig \ref{fig:rlw-overview}) of the DRZ is primarily used for research in order to analyze and evaluate different robot systems in various test scenarios. From this, it is to be derived in which operational areas the robots can support the emergency forces in a meaningful way. In addition to research, the RobLW is used by the FwDO in real operations to test the researched results in practice and to draw conclusions as quickly as possible so that the platforms can be further optimized (Figure \ref{fig:rlw-overview}).

The RobLW is 7m long, 2m wide and 3m high and is divided into 3 sections. 
The front area is used as a command room to control the vehicle and to communicate with other emergency forces, as well as the control center. Therefore, the vehicle has an integrated radio. In addition, it is possible to connect to the vehicle's network with a separate computer. In the rear of the vehicle there are various peripheral devices and components to be able to react autonomously to as many scenarios as possible. In addition to the auxiliary equipment, the rear area provides a storage area for ground robots and UAVs to transport them to the scene. The middle area represents the operations center. From here, the robots and UAVs are controlled by trained personnel and supported by appropriate assistance systems. For this purpose, the RobLW operations center is equipped with hardware for two workstations and a powerful server. Each workstation has a monitor and an Intel NUC with the associated peripherals. The Intel NUCs are used exclusively for the visualization of camera data or for the visual representation of 3D environments. The complex calculations of three-dimensional environments or the inference of artificial neural networks are executed on the server and then shared over the network. The server has an AMD Ryzen Threadripper processor with 64 cores, 64 GB of RAM and an Nvidia Geforce RTX 2080 with 8 GB of dedicated memory. To ensure a successful network, a TDT router is used, which primarily serves as a DHCP server and provides access to the Internet via the Mas cellular network. This router can also be used as a DHCP client if the network of another mission vehicle is to be used. Through the Internet connection it is possible to download local mission maps over the Internet or send current data to the command center to provide a better overview of the current situation. In addition, the router creates its own Wi-Fi network. With this it is possible to view the 3D models and other results, not only inside the car but also outside the car.
Since the server is mainly responsible for processing the various data from the robots, it contains the primary software to support the firefighters. In order to get a better overview of the situation, the server has a developed situational awareness system to which robots can register via ROS messages. After the robots are registered, they are displayed geo-referenced in the situation picture system. In addition, the provided live streams of the robots are displayed in the situation picture system. This offers the task forces the possibility to explore different areas by robots and to analyze them on site at the same time. In addition to the situation picture system, the server has a panorama viewer and the WebODM\footnote{Drone Mapping Software: \url{www.opendronemap.org/webodm/}}. The Pano-Viewer is a web-based tool that displays 360° panoramas in an interactive 360° panorama in order to switch back and forth between the individual panoramas. This tool can be used to filter out detailed information in the high resolution panoramas. The WebODM is also a web-based tool that can be used to create a three-dimensional representation of the environment using Structure from motion, Multi View Stereo and suitable images. If the image data is georeferenced, a scaled point cloud of the environment is created. In addition, the tool creates an orthomap of the current area. If the three-dimensional information and the orthomap are combined, depth profiles of the surroundings can be generated to determine the rise or fall of water in flood areas.

\subsubsection{UAV}

\begin{figure}
\centering
\includegraphics[width=0.45\textwidth]{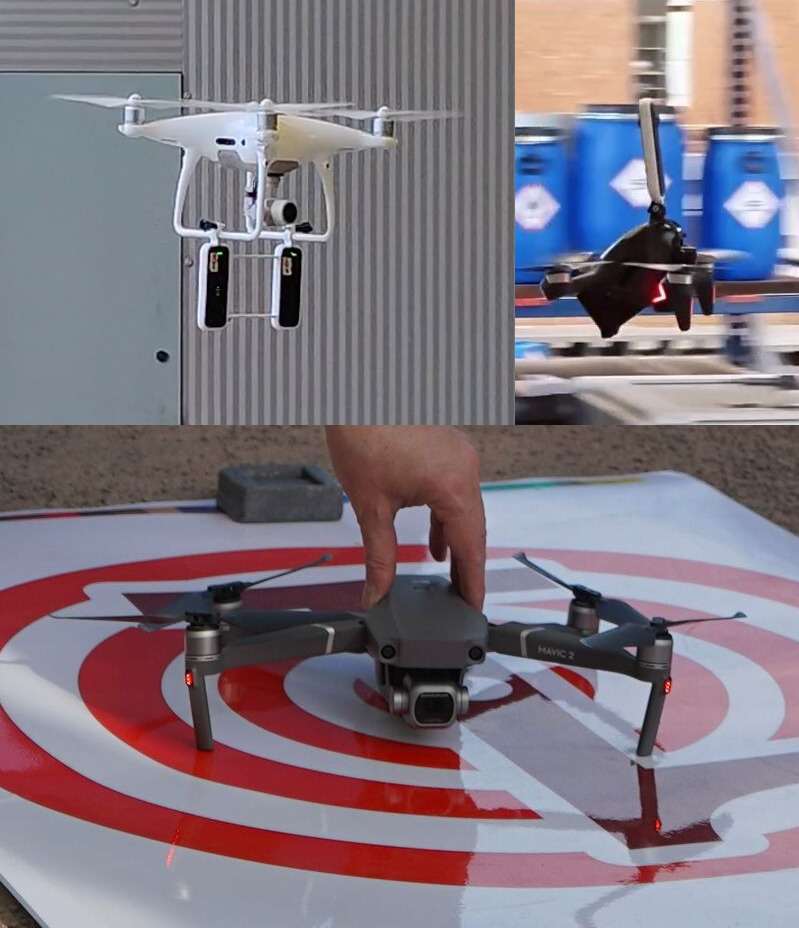}
\caption{The used UAVS: Phantom 4 with two additional stereo 360 cameras, DJI FPV with one additional 360° camera and a DJI mavic zoom.}
\label{fig:uavs}
\end{figure}

Due to the large area, the use of UAVs is mandatory, because they provide a basic overview as quickly as possible. In addition, the risk of losing the drone if the ground collapses or supports a house is much lower than with a ground robot.  Therefore, several drones were taken into the field (Figure \ref{fig:uavs}). The DJI Mavic Pro (MP2) was used to capture detailed footage of the environments. This drone has a camera resolution of 5472x3648 a maximum video bit rate of 100 Mbps. In addition to the MP2, the DJI Mavic 2 Zoom (MZ2) was used, this has a lower resolution than the MP2 at 4000x3000 pixels, however the MZ2 has a larger field of view than the MP2 with a camera aperture of 83°. Since it may also be necessary to fly into buildings, a drone with as large a field of view as possible was needed. For this purpose, the DJI FPV drone with the DJI goggles was used, because these have an aperture angle of 150°. However, this has an even worse camera resolution than the MZ2 and is therefore only used for flying in or near buildings. The DJI Phantom 4 (P4P) was chosen for its stable flight characteristics. This allows one or two panoramic cameras to be attached to the UAV without it exhibiting unstable flight behavior.

\subsubsection{Cameras}
To get as much information as possible from a picture in a short time, panorama cameras are used. These are attached to specially developed mounts, for the P4P as well as for the FPV. The camera used was an Insta 360 One X (Figure \ref{fig:uavs}). This has a video resolution of 5760x2880 (15.8MP) at 30 FPS. This allows the camera to provide a lot of high-resolution image information of the environment while the UAV travels a short distance.

\subsection{Overview}

\begin{figure}
\centering
\includegraphics[width=0.45\textwidth]{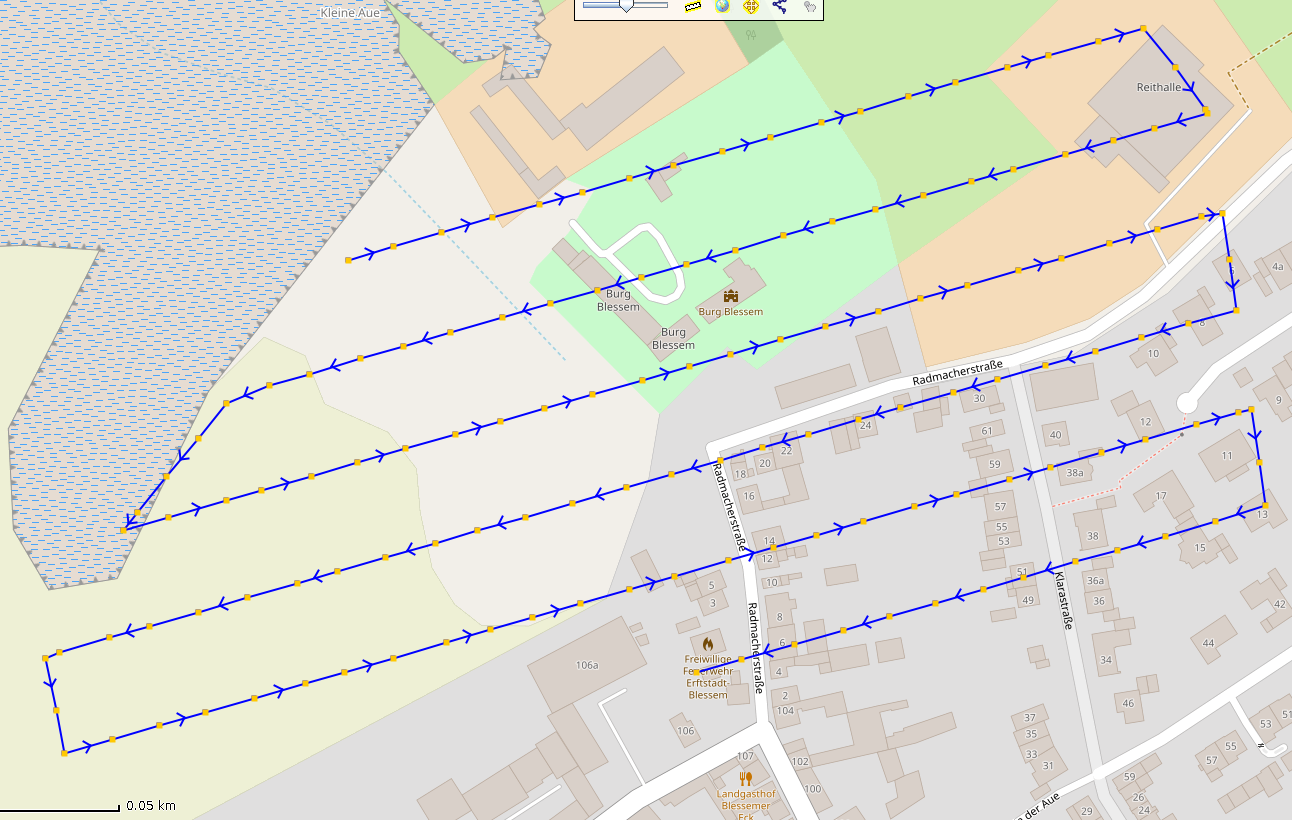}
\caption{Example of a pre-planned flight trajectory which is then processed autonomously by the UAV.}
\label{fig:flight-trajectorie}
\end{figure}

\begin{figure*}[!t]
\centering
\includegraphics[width=0.95\textwidth]{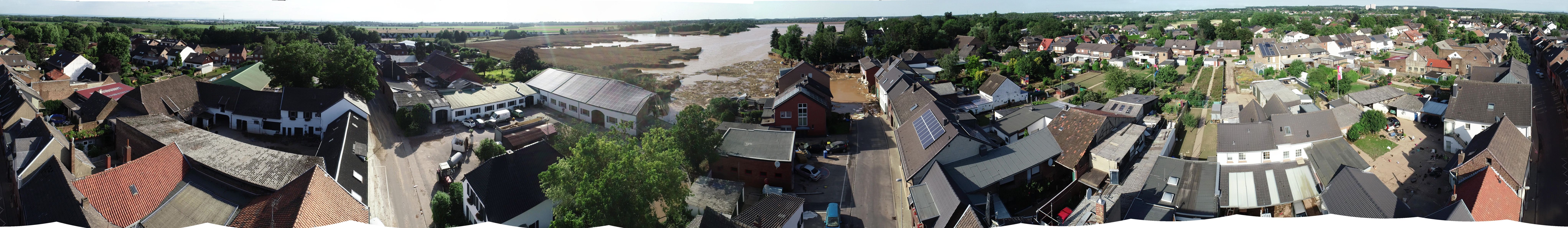}
\caption{360° panorama of the city and the demolition edge in Blessem / NRW}
\label{fig:pano}
\end{figure*}

\begin{figure}
\centering
\includegraphics[width=0.48\textwidth]{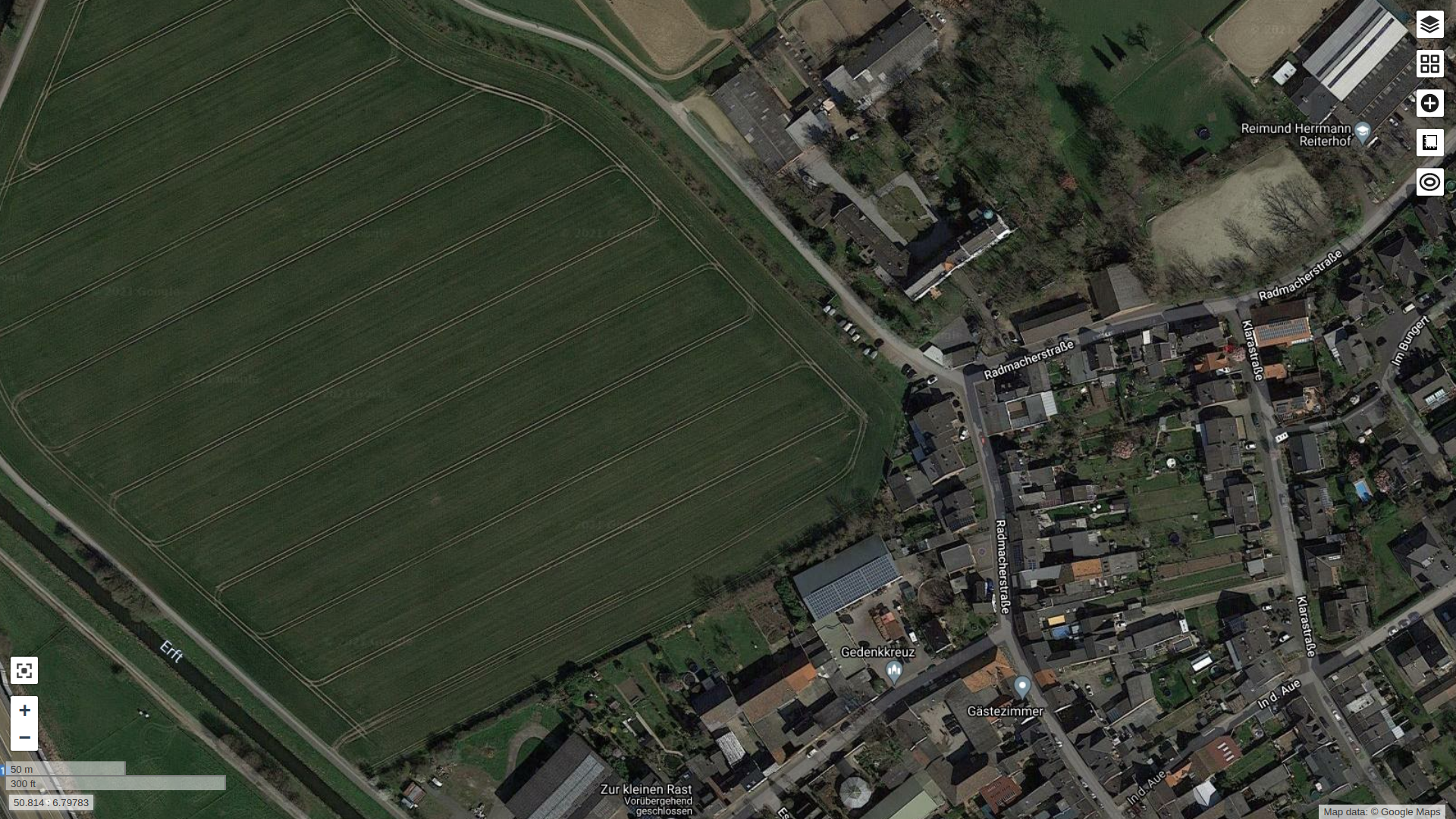}
\includegraphics[width=0.48\textwidth]{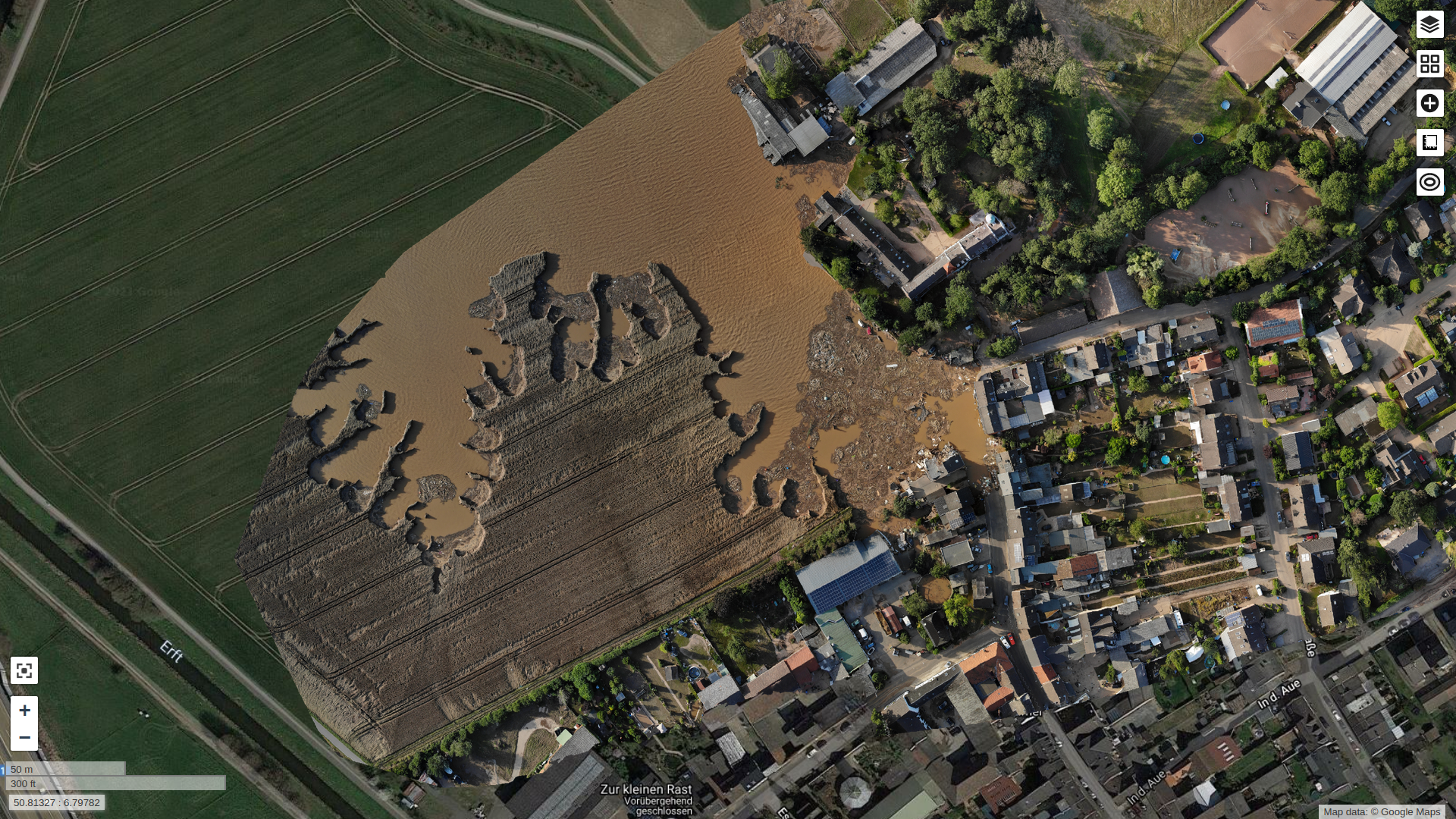}
\caption{Comparison of orthophotos to estimate the degree of destruction. Above before the disaster and below after the disaster. The orthophoto has been calculated from 199 single images and can be overlaid live in RobLW with a slider over google maps.}
\label{fig:orthomaps}
\end{figure}

\begin{figure}
\centering
\includegraphics[width=0.48\textwidth]{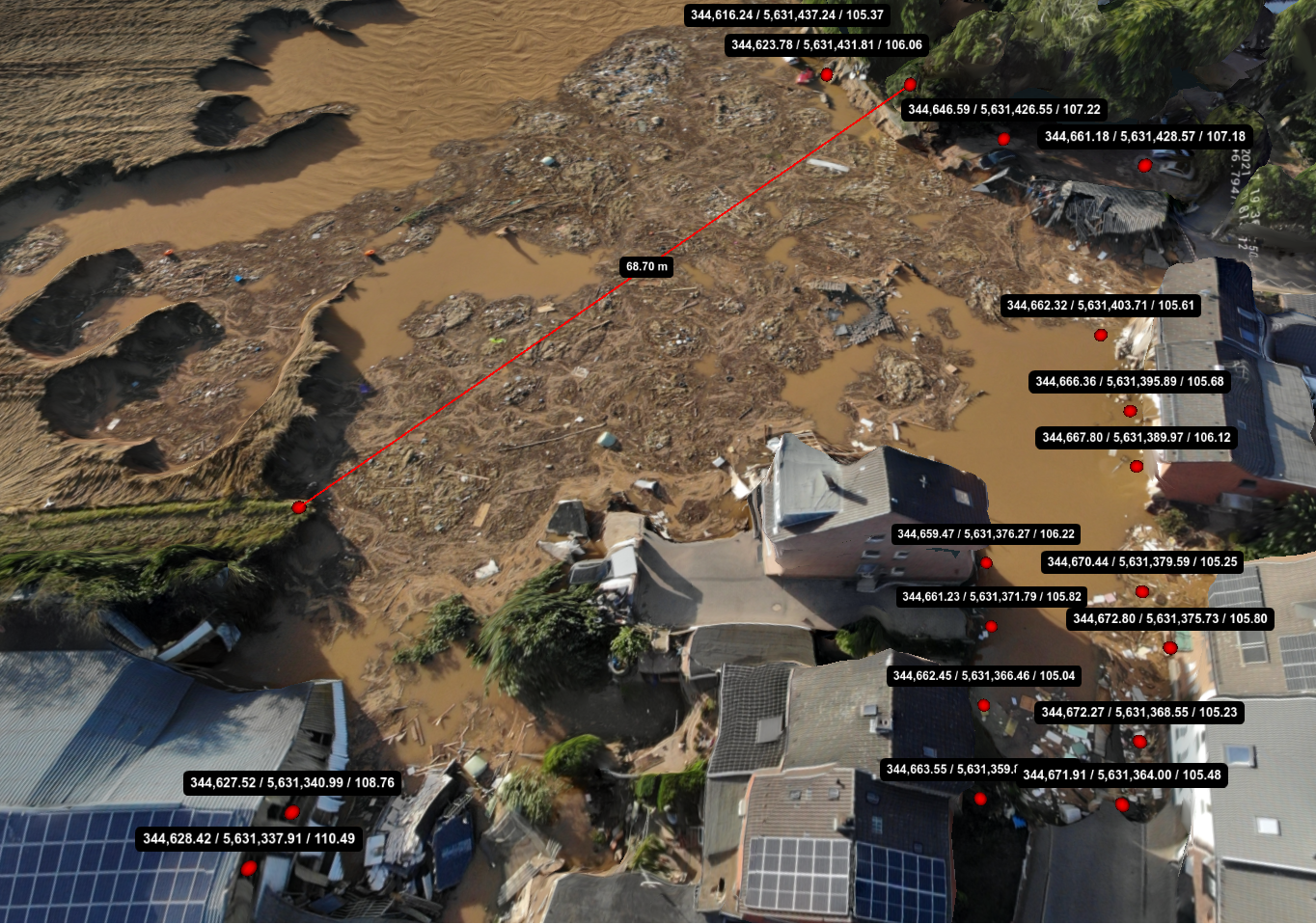}
\caption{Measurements and planning in 3D model. The selected points are inspected with the FPV UAV according to the specification.}
\label{fig:plan}
\end{figure}
Due to the flood disaster, many people were missing, so emergency forces were to search the surrounding area and the houses. However, since the area was destroyed over a large area, planning with existing information was not possible. Therefore, a meander flight was planned with the MP2 using GPS coordinates (Figure \ref{fig:flight-trajectorie}). In the process, 199 images were created, which were then used for calculation on the server in RobLW. From these georeferenced images a georeferenced orthophoto was created. This can be overlaid with Google Maps to assess the current extent of destruction (Figure \ref{fig:orthomaps}). In addition, particularly critical locations could be taken from the orthophoto, such as the break-off edge, because near this edge the ground was threatening to break away due to the receding water. In addition to the meander flight, a 360° panorama was created by taking individual photographs in order to obtain an overview as quickly as possible (Figure \ref{fig:pano}).
Since this information was sufficient for a rough planning, but the risk for all involved should be further minimized, 3D models of the environment were created by means of Structure from motion and Multi View Stereo, in order to be able to guarantee a more exact planning. Figure \ref{fig:plan} shows a scaled model. It shows which areas were particularly risky and should be explored by a drone.

\subsection{Detailed inspection}

For the detailed investigation, the FPV drone from DJI was used, as this drone has the largest opening angle. The large opening angle makes it possible to explore buildings or other narrow areas without an additional drone, such as in Berlin, and to leave this environment again with little risk. 

\begin{figure}
\centering
\includegraphics[width=0.42\textwidth]{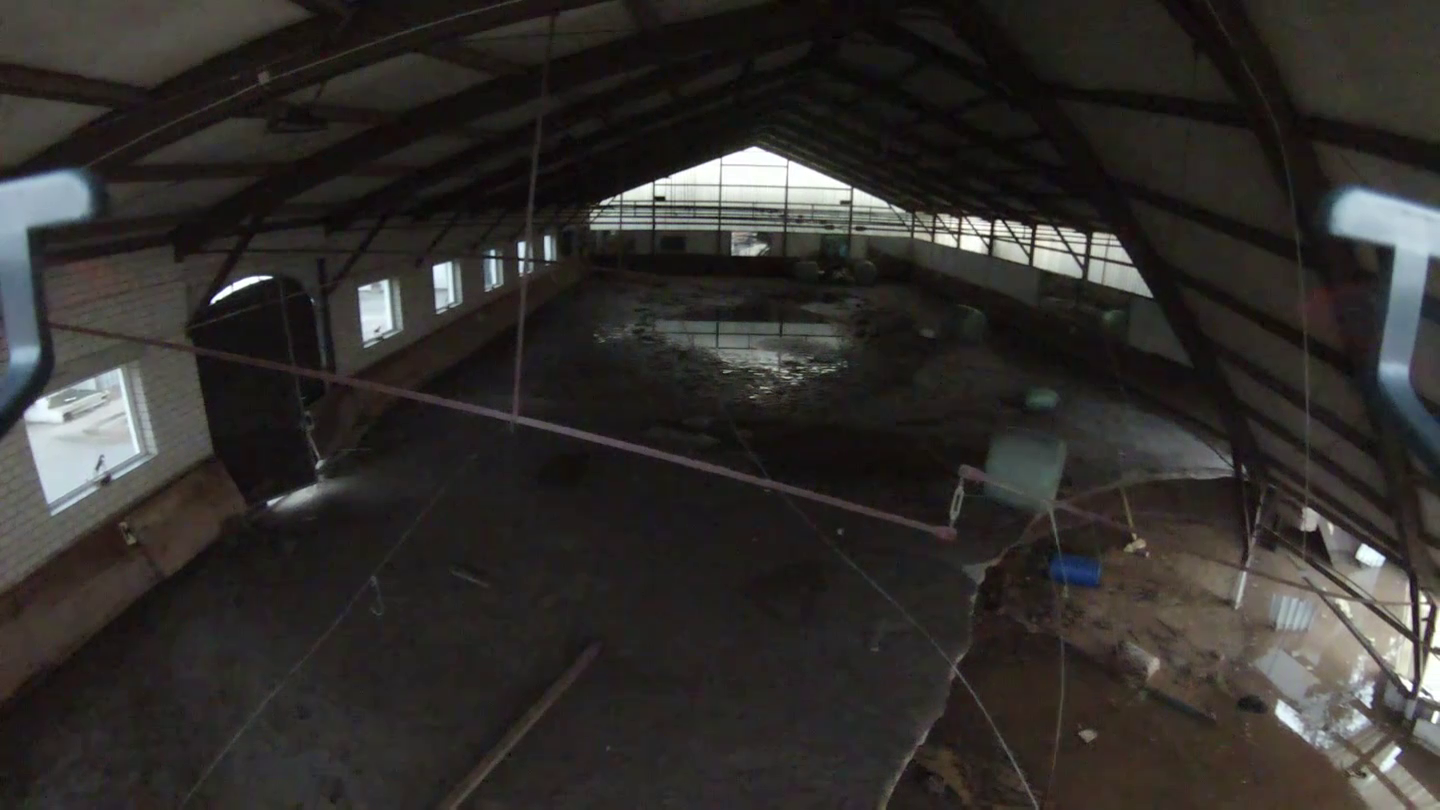}
\includegraphics[width=0.42\textwidth]{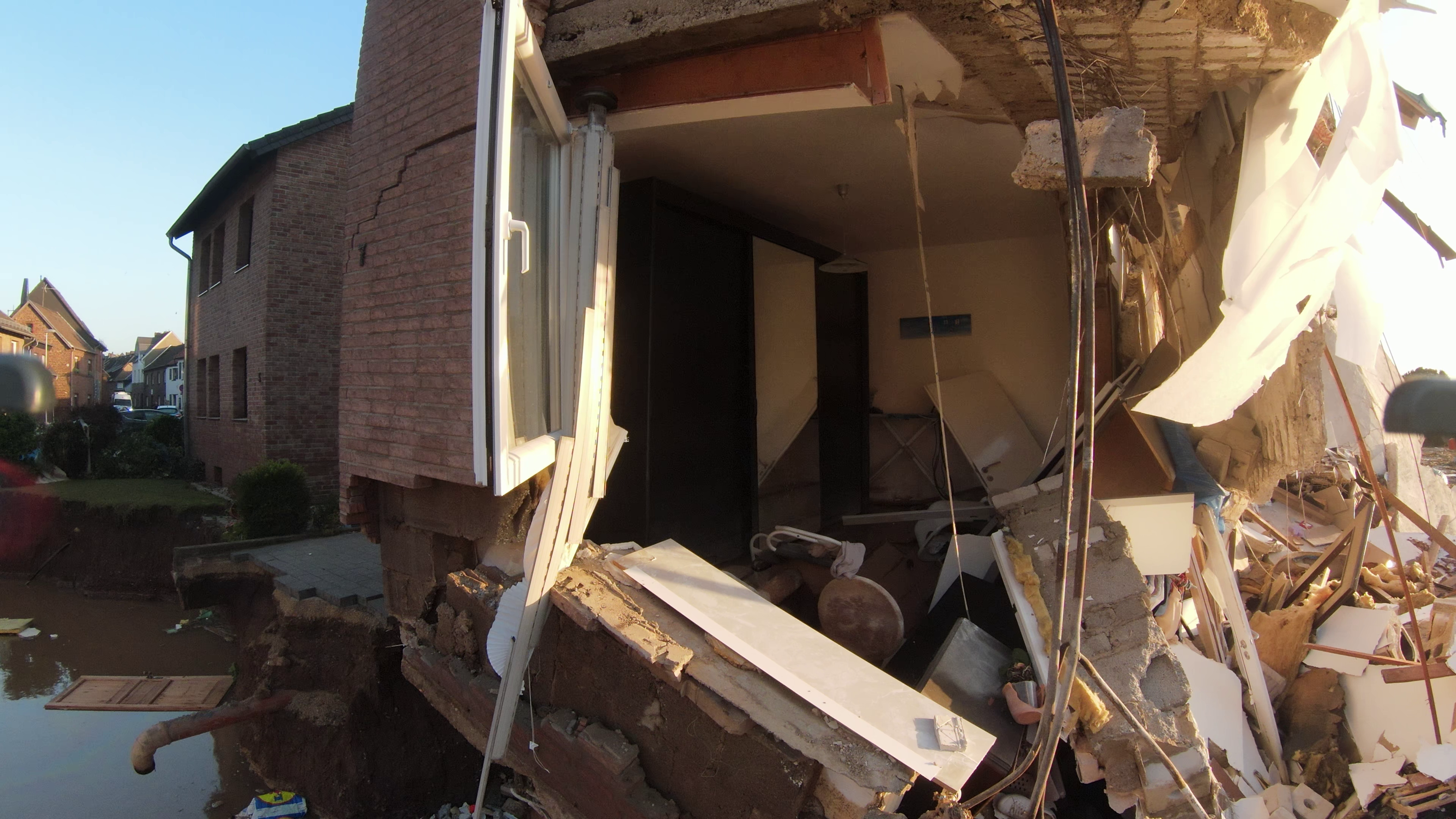}
\includegraphics[width=0.42\textwidth]{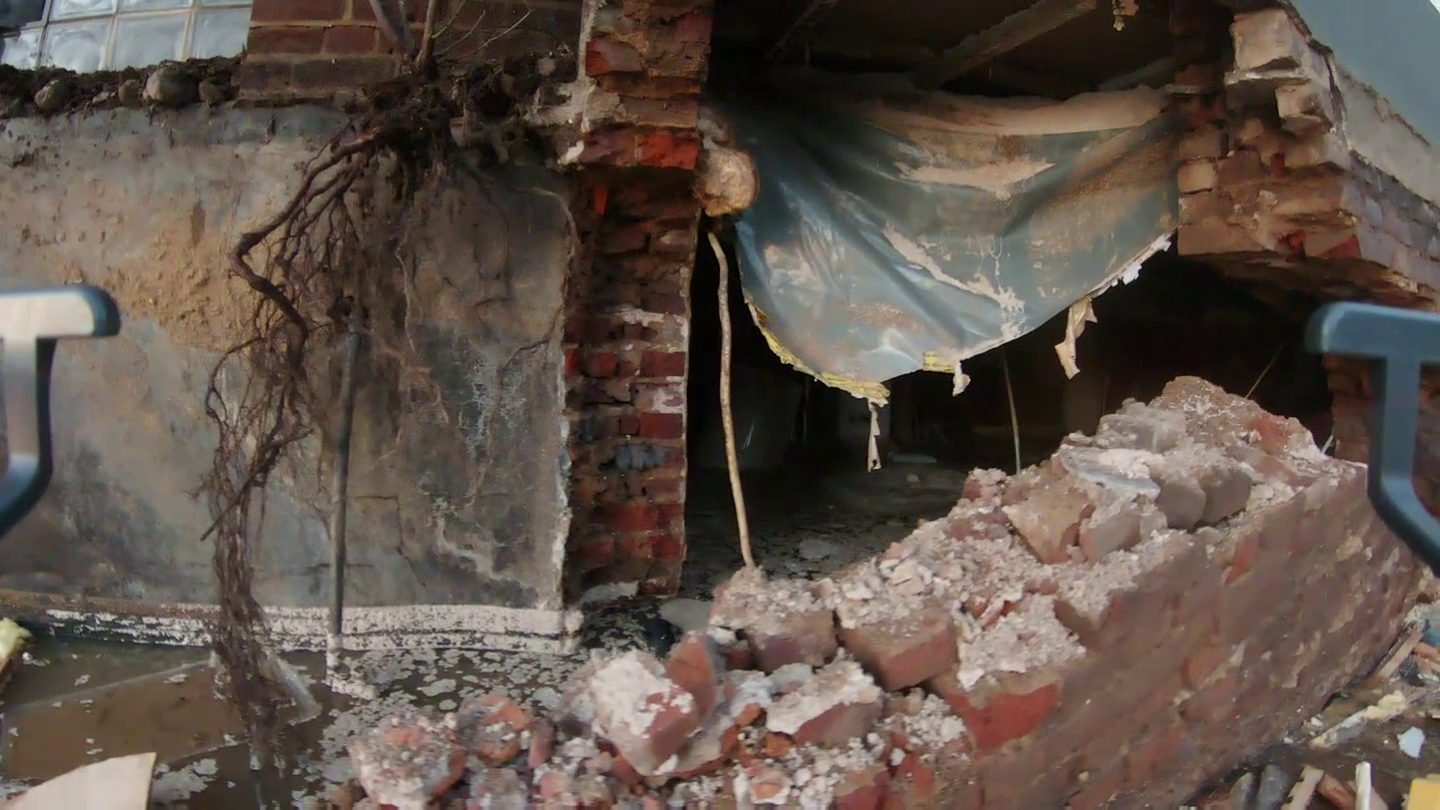}
\includegraphics[width=0.42\textwidth]{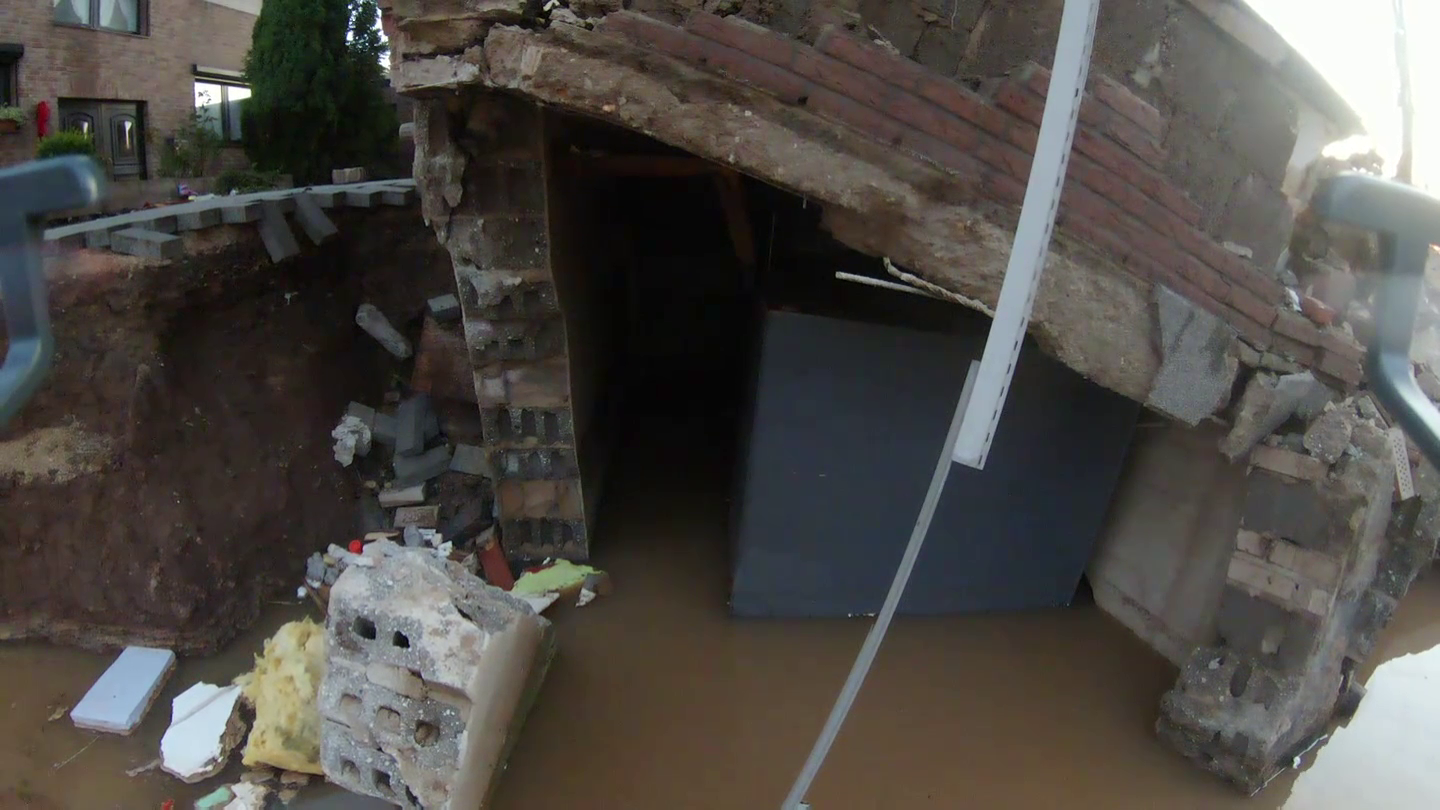}
\caption{Sample images of some of the points marked in Figure 7 from the detailed aerial survey taken with the FPV UAV.}
\label{fig:collections_details}
\end{figure}

During the detailed exploration, an attempt was made to explore areas that were difficult to access or that posed a risk. These were flown through destroyed areas, such as a broken away entrance or a partially collapsed house (Figure \ref{fig:collections_details}). To increase the chance of finding missing persons and to support the emergency services, an AI detector based on semantic segmentation was used to detect people on the image information.

\begin{figure}[!t]
\centering
\includegraphics[width=0.43\textwidth]{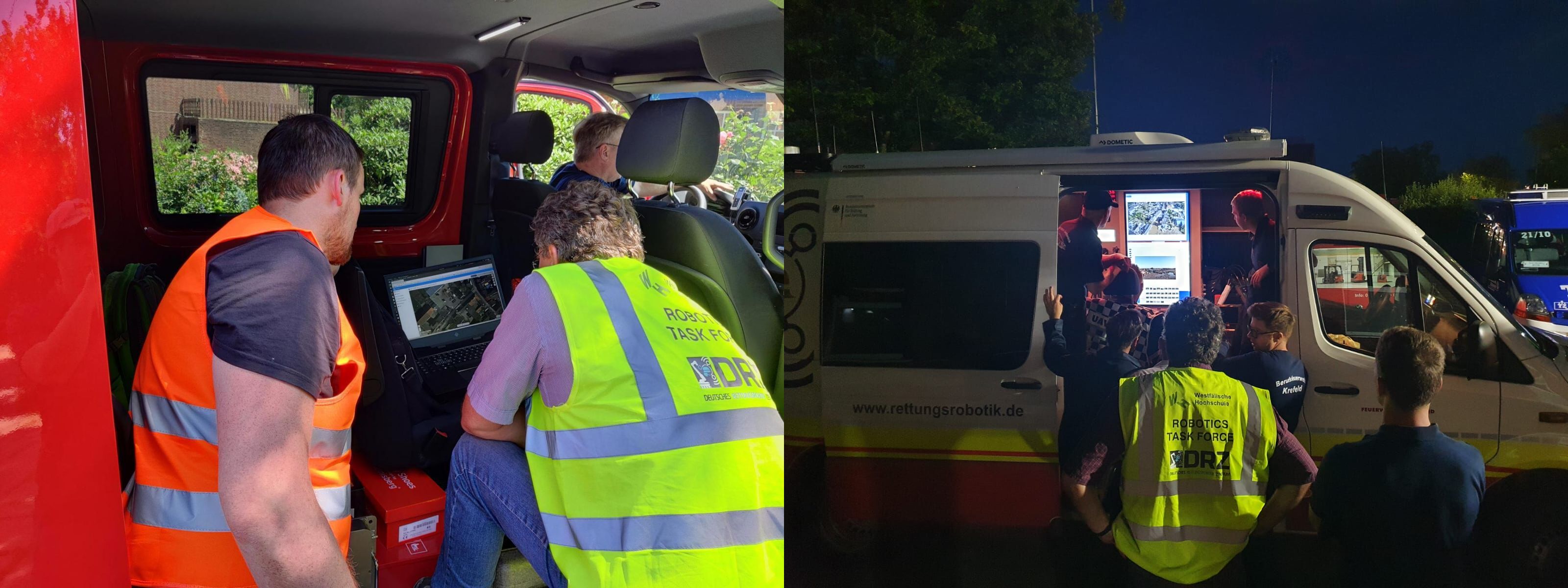}
\caption{Left: Due to the high demand for computing power, laptops with docker containers of the mapping software were distributed to additional vehicles and used for mission planning. Right: Nightly presentation of the mission images collected throughout the day and the calculated results on the two monitors in RobLW. }
\label{fig:besprechung_nacht}
\end{figure}

Due to the partly heavy destruction of the buildings, it was not possible to perform an optimal exploratory search in all areas. Therefore, all access points to the areas that had not been explored were examined more closely and assigned a risk potential by corresponding experts. Based on this information, the task forces were able to plan a further procedure to investigate the areas (Figure \ref{fig:besprechung_nacht}).

\section{Outcomes}
\subsection{3D overview image}

\begin{figure}
\centering
\includegraphics[width=0.43\textwidth]{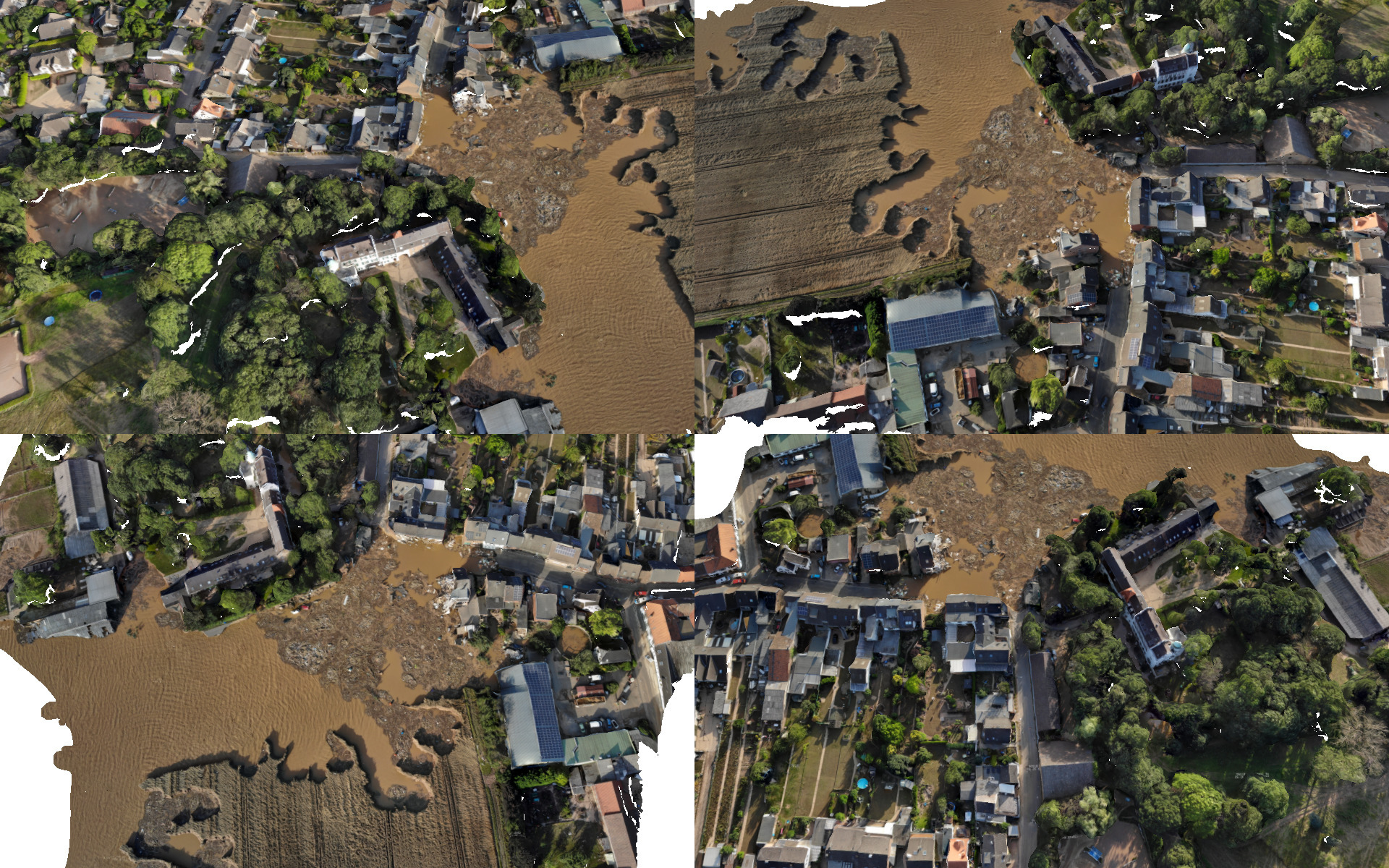}
\caption{Four different views of the 3D model calculated by Structure from motion and MultiView Stereo within 15 minutes after the landing of the UAVs..}
\label{fig:3dmodels}
\end{figure}

Through a variety of 3D models (Figure \ref{fig:3dmodels}), different areas could be represented to different degrees. This generally ensured that the emergency forces had to look at different models, but that sufficient information was available. However, it was found that the data from the meander flights provided sufficient information for the responders. Using the data from the detailed flights, 3D models could be created, but the information was partially unusable. 
Therefore, it can be seen that 3D models based on image information provide a good overall view, but have difficulties in reconstructing collapsed buildings.

\subsection{Altitude profiles}

\begin{figure}
\centering
\includegraphics[width=0.42\textwidth]{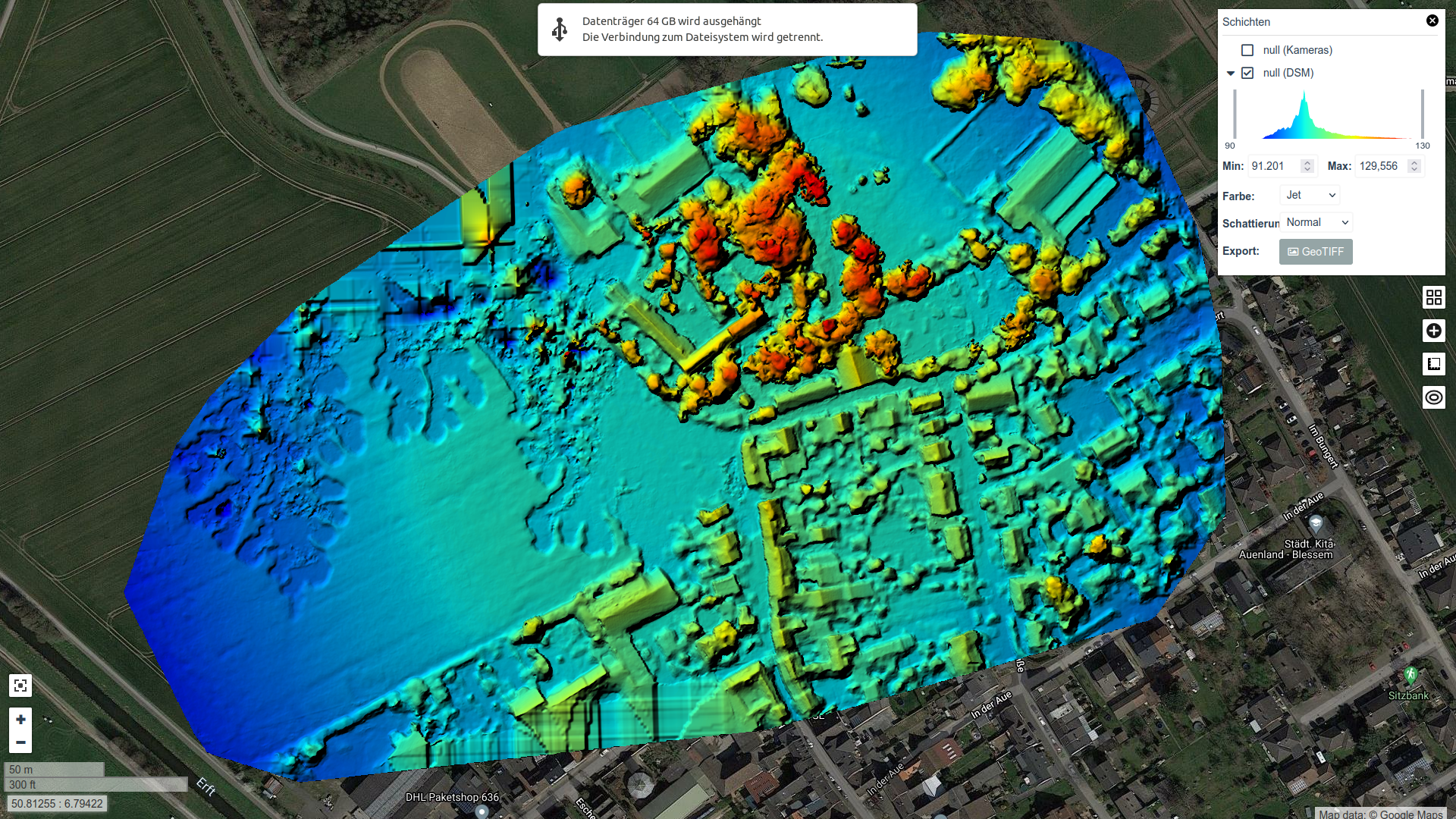}
\includegraphics[width=0.42\textwidth]{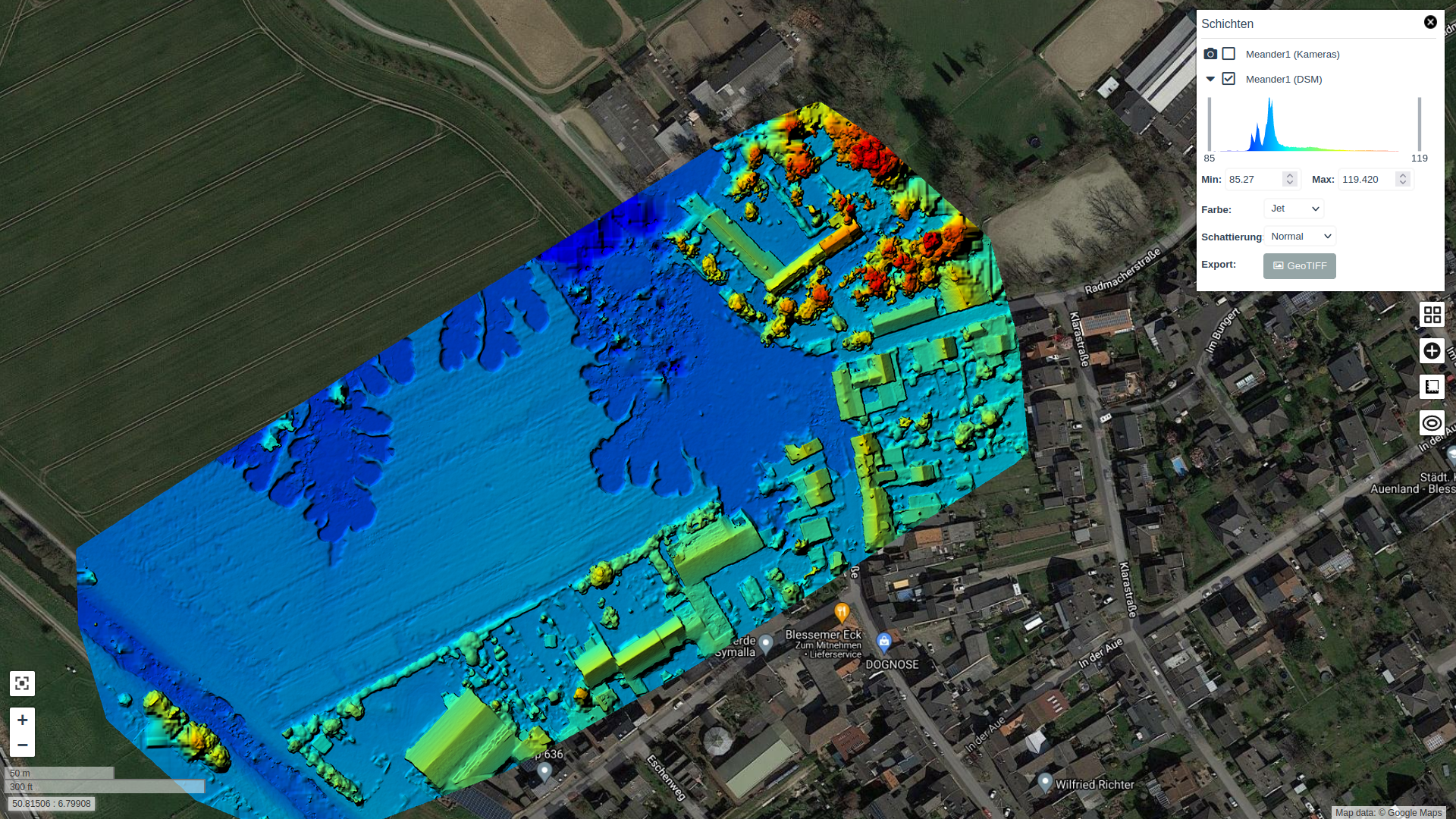}
\caption{Elevation profiles of consecutive days. The comparison of the two elevation profiles shows that the water has sunk by about 40 cm within one day, threatening further demolitions due to the lack of water backpressure. Based on the information, the emergency forces were pulled back by 100 meters from the demolition site.}
\label{fig:tiefenprofile}
\end{figure}

Through the daily meander flight and the associated creation of an orthophoto, up-to-date information could be used for planning. However, it became apparent that the combination of depth information and an orthophoto produced much more useful information. The resulting depth profiles show the depth differences. Figure \ref{fig:tiefenprofile} shows depth profiles from two consecutive days. From these images, it can be deduced that the water has receded significantly within a day, significantly increasing the danger to responders near the edge of the breach. Concluding from this, such flights must be made not daily but hourly in order to determine the current recession of the water and to reduce the risk for all involved to a minimum.


\section{CONCLUSIONS}
Climate change will lead to more and more extreme weather events in the coming years. In order to support society, it is necessary to make current research results and algorithms from the computer science and robotics laboratories available to society as quickly as possible, in particular to the emergency services. This is exactly where this article comes in. It shows how current technologies such as UAVs and algorithms (struture from motion / SLAM, multiView Stereo / 3D pointclouds) from robotics (labs) were used during the flood disaster in July 2021 in Erftstadt / Blessem, but also where further work is needed. 
After the flood there were large areas flooded, buildings washed out and / or collapsed. It has been shown that the use of autonomous meander flights of the UAVs provide optimal images of the surrounding area. These are computed within a few minutes on site with the help of the RobLW independently of the Internet and power supply to high-quality 3D models. These are then used for mission planning and detailed reconnaissance. These detailed reconnaissances are carried out with small FPV UAVs, whereby a large risk of collision and loss is assumed here. The used DJI FPV had only been available for 3 months. The flight data were automatically processed into reports and after use are fed into global GIS systems where they can be compared with satellite images. Needless to say a lot of work remains to be done. AI support for damage processing and assessment is urgently needed as well as improved (partly autonomous) flight support especially during detailed inspection with UAVs.

\section*{Acknowledgment}
This work was founded by the Federal Ministry of Education and Research (BMBF) under grant number 13N14860 (A-DRZ https://rettungsrobotik.de/).






\bibliographystyle{IEEEtran} 
\bibliography{IEEEabrv,literatur} 

\end{document}